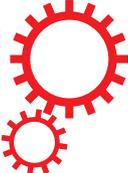

OPEN

# Estimating the intrinsic dimension of datasets by a minimal neighborhood information

Elena Facco, Maria d'Errico, Alex Rodriguez & Alessandro Laio

Analyzing large volumes of high-dimensional data is an issue of fundamental importance in data science, molecular simulations and beyond. Several approaches work on the assumption that the important content of a dataset belongs to a manifold whose *Intrinsic Dimension* (ID) is much lower than the crude large number of coordinates. Such manifold is generally twisted and curved; in addition points on it will be non-uniformly distributed: two factors that make the identification of the ID and its exploitation really hard. Here we propose a new ID estimator using only the distance of the first and the second nearest neighbor of each point in the sample. This extreme minimality enables us to reduce the effects of curvature, of density variation, and the resulting computational cost. The ID estimator is theoretically exact in uniformly distributed datasets, and provides consistent measures in general. When used in combination with block analysis, it allows discriminating the relevant dimensions as a function of the block size. This allows estimating the ID even when the data lie on a manifold perturbed by a high-dimensional noise, a situation often encountered in real world data sets. We demonstrate the usefulness of the approach on molecular simulations and image analysis.

The latest developments in hardware and software technology have led to a drastic rise in data availability: as claimed in ref.[1], "The era of big data has come beyond all doubt". This is triggering the development of more and more advanced approaches aimed at analyzing, classifying, structuring and simplifying this bunch of information in order to make it meaningful and usable. Data are usually represented by high-dimensional feature vectors, but in many cases they could be in principle embedded in lower dimensional spaces without any information loss: this dimensionality reduction is often necessary for algorithmic strategies to work in practice. In recent years, a variety of techniques have been proposed to reduce the dimensionality of data[2]; to this purpose, a fundamental question is: which is the minimum number of variables needed to accurately describe the important features of a system? Such number is known as Intrinsic Dimension (ID). Information about the ID of a dataset is relevant in many different contexts, for instance in molecular simulations, where often a dimensionality reduction is required[3], or in bioinformatics[4], or in image analysis where the ID is a suitable descriptor to distinguish between different kinds of image structures[5]. Estimating the ID can be a hard task; data are almost invariably characterized by density variations, and this makes the estimate of the ID entangled with the estimate of the density. Moreover, data often lie on a topologically complex curved manifold. A further issue arising in high dimensions is that data behave in an extremely counterintuive way due to the so called curse of dimensionality[4]: the smallest sampled distance increases with the ID, and nearly all of the space is spread "far away" from every point. An even more subtle problem is that in real datasets not all the dimensions have the same importance; some of them identify directions in which the features of the dataset change remarkably, while others are characterized by small variations that can be irrelevant for the analysis and can be labeled as "noise". Consider for example a sample of configurations explored during a molecular dynamics run at finite temperature of a complex biomolecule. In the absence of constraints on the bond length, the intrinsic dimension of the hypersurface explored in this dynamics is 3*N*, where *N* is the number of atoms. A well defined estimator should in principle provide this number when infinitely many configurations are sampled. However, this asymptotic estimate is clearly irrelevant for practical purposes. Indeed most of the 3*N* possible directions are highly restrained, due for example to steric clashes between neighboring atoms, and those in which the system can move by a significant amount are normally much fewer. A practically meaningful estimate of the ID should provide the number of these *soft* directions.

SISSA International School for Advanced studies, department of Molecular and Statistical Biophysics, Trieste, 34136, Italy. Correspondence and requests for materials should be addressed to A.L. (email: laio@sissa.it)

Different approaches have been developed to cope with the ID estimation problem. *Projection* techniques look for the best subspace to project the data by minimizing a projection error[6] or by preserving pairwise distances[7–9] or local connectivity[10]. Another point of view is given by *fractal* methods, for instance[11]: based on the idea that the volume of a *d*-dimensional ball of radius *r* scales as $r^d$, they count the number of points within a neighborhood of radius *r* and estimate the rate of growth of this number; these methods in general have the fundamental limitation that in order to obtain an accurate estimation the number of points in the dataset has to be exponentially high with respect to the dimension. In ref.[4] this difficulty is addressed, and a multiscaling analysis is discussed. Also in refs[12,13] an estimate of the dimension is provided that depends on the scale. The fractal dimension can also be inferred from the probability distribution of the first neighbour[14]. Finally, *Nearest Neighbors-Based* ID estimators describe data neighborhoods distributions as functions of the intrinsic dimension *d*, usually assuming that close points are uniformly drawn from small enough *d*-dimensional hyperspheres (MLE[15], DANCO[16]).

Building on the premises in ref.[15], we here introduce TWO-NN, a new ID-estimator that employs only the distances to the first two nearest neighbors of each point: this minimal choice for the neighborhood size allows to lower the influence of dataset inhomogeneities in the estimation process. If the density is approximately constant on the lengthscale defined by the typical distance to the second neighbor it is possible to compute the distribution and the cumulative distribution of the ratio of the second distance to the first one, and it turns out that they are functions of the intrinsic dimension *d* but not of the density; at this point an equation is obtained that links the theoretic cumulate *F* to *d*, and by approximating *F* with the empirical cumulate obtained on the dataset we are able to estimate the intrinsic dimension.

We further discuss the applicability of the method in the case of datasets characterized by non-uniform density and curvature. First of all we show the asymptotic convergence of the estimated ID to the correct one as the number of points in the sample increases; then we analyze the behavior of TWO-NN in the case of datasets displaying density variations and curvatures, up to dimension 20. We address the problem of multiscaling, proposing a technique to detect the number of meaningful dimensions in the case of noise; we demonstrate the accuracy of the procedure analyzing two datasets of images, the Isomap face dataset and a dataset extracted from the MNIST database[9]; finally we investigate the intrinsic dimension of the configurational space explored in a molecular dynamics simulation of the RNA trinucleotide AAA[17], obtaining comparable results under the choice of two different commonly used distances between configurations.

## Results

Let *i* be a point in the dataset, and consider the list of its first *k* nearest neighbors; let $r_1, r_2, \ldots, r_k$ be a sorted list of their distances from *i*. Thus, $r_1$ is the distance between *i* and its nearest neighbor, $r_2$ is the distance with its second nearest neighbor and so on; in this definition we conventionally set $r_0 = 0$.

The volume of the hypersferical shell enclosed between two successive neighbors $l-1$ and *l* is given by

$$\Delta v_l = \omega_d (r_l^d - r_{l-1}^d), \tag{1}$$

where *d* is the dimensionality of the space in which the points are embedded and $\omega_d$ is the volume of the *d*-sphere with unitary radius. It can be proved (see SI for a derivation) that, if the density is constant around point *i*, all the $\Delta v_l$ are independently drawn from an exponential distribution with rate equal to the density $\rho$:

$$P(\Delta v_l \in [v, v+dv]) = \rho e^{-\rho v} dv. \tag{2}$$

Consider two shells $\Delta v_1$ and $\Delta v_2$, and let *R* be the quantity $\frac{\Delta v_i}{\Delta v_j}$; the previous considerations allow us, in the case of constant density, to compute exactly the probability distribution (pdf) of *R*:

$$P(R \in [\overline{R}, \overline{R} + d\overline{R}])$$
$$= \int_0^\infty dv_i \int_0^\infty dv_j \rho^2 e^{-\rho(v_i+v_j)} 1_{\left\{\frac{v_j}{v_i} \in [\overline{R}, \overline{R}+d\overline{R}]\right\}}$$
$$= d\overline{R} \frac{1}{(1+\overline{R})^2},$$

where 1 represents the indicator function. Dividing by $d\overline{R}$ we obtain the pdf for *R*:

$$g(R) = \frac{1}{(1+R)^2}. \tag{3}$$

The pdf does not depend explicitly on the dimensionality *d*, which appears only in the definition of *R*. In order to work with a cdf depending explicitly on *d* we define quantity $\mu \doteq \frac{r_2}{r_1} \in [1, +\infty)$. *R* and $\mu$ are related by equality

$$R = \mu^d - 1. \tag{4}$$

This equation allows to find an explicit formula for the distribution of $\mu$:

$$f(\mu) = d\mu^{-d-1} 1_{[1,+\infty]}(\mu), \tag{5}$$

while the cumulative distribution (cdf) is obtained by integration:

$$F(\mu) = (1 - \mu^{-d})1_{[1,+\infty]}(\mu). \tag{6}$$

Functions $f$ and $F$ are independent of the local density, but depend explicitly on the intrinsic dimension $d$.

**A Two Nearest Neighbors estimator for intrinsic dimension.** The derivation presented above leads to a simple observation: the value of the intrinsic dimension $d$ can be estimated through the following equation

$$\frac{log(1 - F(\mu))}{log(\mu)} = d. \tag{7}$$

Remarkably the density $\rho$ does not appear in this equation, since the cdf $F$ is independent of $\rho$. This is an innovation with respect to, for instance ref.[14], where the dimension estimation is susceptible to density variations. If we consider the set $S \subset \mathbb{R}^2, S \doteq \{(log(\mu), -log(1 - F(\mu)))\}$, equation 7 claims that in theory $S$ is contained in a straight line $l \doteq \{(x, y) \mid y = d * x\}$ passing through the origin and having slope equal to $d$. In practice $F(\mu)$ is estimated empirically from a finite number of points; as a consequence, the left term in equation 7 will be different for different data points, and the set $S$ will only lie around $l$. This line of reasoning naturally suggests an algorithm to estimate the intrinsic dimension of a dataset:

1. Compute the pairwise distances for each point in the dataset $i = 1, ..., N$.
2. For each point i find the two shortest distances $r_1$ and $r_2$.
3. For each point i compute $\mu_i = \frac{r_2}{r_1}$.
4. Compute the empirical cumulate $F^{emp}(\mu)$ by sorting the values of $\mu$ in an ascending order through a permutation $\sigma$, then define $F^{emp}(\mu_{\sigma(i)}) \doteq \frac{i}{N}$.
5. Fit the points of the plane given by coordinates $\{(log(\mu_i), -log(1 - F^{emp}(\mu_i))) \mid i = 1, ..., N\}$ with a straight line passing through the origin.

Even if the results above are derived in the case of a uniform distribution of points in equations (5) and (7) there is no dependence on the density $\rho$; as a consequence from the point of view of the algorithm we can a posteriori relax our hypothesis: we require the dataset to be only *locally* uniform in density, where locally means in the range of the second neighbor. From a theoretical point of view, this condition is satisfied in the limit of $N$ going to infinity. By performing numerical experiments on datasets in which the density is non-uniform we show empirically that even for a finite number of points the estimation is reasonably insensitive to density variations. The requirement of local uniformity only in the range of the second neighbor is an advantage with respect to competing approaches where local uniformity is required at larger distances.

**Benchmark.** In Fig. 1 we plot $-log(1 - F^{emp}(\mu_i))$ as a function of $log(\mu_i)$ for three exemplar datasets containing 2500 points: a dataset drawn from a uniform distribution on a hypercube in dimension $d = 14$, analyzed with periodic boundary conditions (pbc), a dataset drawn from a uniform distribution on a Swiss Roll embedded in a three-dimensional space, and a Cauchy dataset in $d = 20$. By "Cauchy dataset" we refer to a dataset where the norms of points are distributed according to the pdf $f(x) = \frac{1}{1+x^2}$. The hypercube with pbc is the pdf that best resembles a uniform distribution on a linear space; nevertheless it has to be noticed that the pbc introduce correlations in the distances whenever the typical distance of the second neighbor is comparable with the box size. In the same figure, we draw the straight line passing through the origin and fitting the points $\{(log(\mu_i), -log(1 - F^{emp}(\mu_i))) \mid i = 1, ..., N\}$. The slope of this line is denoted in the following by $\hat{d}$. According to the TWO-NN estimator, the value of the ID for the uniform hypercube is $\hat{d} = 14.09$, a measure that is consistent with the ground truth values. For the Swiss Roll the ID estimated by TWO-NN is 2.01. This value corresponds to the dimension of a hyperplane tangent to the Swiss Roll: in fact, by employing only the first two neighbors of each point, the TWO-NN estimator is sensible to the local dimension even if the points are relatively few and are embedded in a curved hypersurface. For the Cauchy dataset, we obtain $\hat{d} = 6.05$, a value sizeably different from the correct one. Indeed, the slope of the fitting line is strongly affected by a few points characterized by a high value of $\mu_i$. In distributions characterized by heavy tails there is a significant probability of having $r_2 \gg r_1$ and a large value of the ratio $\frac{r_2}{r_1}$. This makes the fit unstable. In order to cope with these situations and make the procedure more robust, we discard the 10% of the points characterized by highest values of $\mu$ from the fitting. The slopes of the lines obtained in this manner are 13.91, 2.01 and 22.16 for the hypercube, the Swiss Roll and the Cauchy dataset respectively. Remarkably, the value of the slope is practically unchanged for the hypercube and the Swiss Roll, while it is quite different for the Cauchy dataset; in this case by discarding the last points the measure is closer to the ground truth, the fit is more stable and the overall procedure more reliable. Therefore, from now on we discuss results obtained by fitting the line only on the first 90% of the points. In SI we discuss more in detail the effects of discarding different fractions of data points, and show that the estimate for the dimension is robust with respect to this threshold.

We then tested the asymptotic convergence of $\hat{d}$ when the number of points goes to infinity.

As the number of points drawn from a probability distribution grows the distances to the second neighbor get smaller and the effects of curvature and density variations become negligible. As a consequence, the hypothesis of local uniformity in the range of the second neighbor is more strongly justified and the distribution of $\mu$ approximates better and better the pdf $f$; moreover, as the number of points goes to infinity the empirical cumulate $F^{emp}$ converges to the correct one $F$ almost surely. Hence we expect the estimates obtained by TWO-NN to approach the correct value. In Fig. 2 we analyze the asymptotic behavior of the measure obtained on a uniform hypercube with periodic boundary conditions, a gaussian distribution, a Cauchy dataset, and a uniform distribution on a hypersphere. The Gaussian and the Cauchy datasets are interesting test cases as they display a variation in density, while the hypersphere is a case of a uniform distribution on a curved space. In all the cases the estimated dimension appears to converge to the real one. The convergence is faster at lower dimensions: such behavior is expected since if we fix the number of points and the size of the domain the average distance to the second neighbor is shorter in the case of low dimensions, and the hypothesis of local uniformity is closer to being satisfied. The Cauchy dataset is characterized by high variance in the case of a few points, due to the presence of outliers in the $S$ set even when the 10% of points with higher $\mu$ is discarded. We performed additional tests by comparing the estimates of TWO-NN with those obtained with DANCo[16], one of the best state-of-the-art methods according to ref.[2]. For a detailed description of the results see SI.

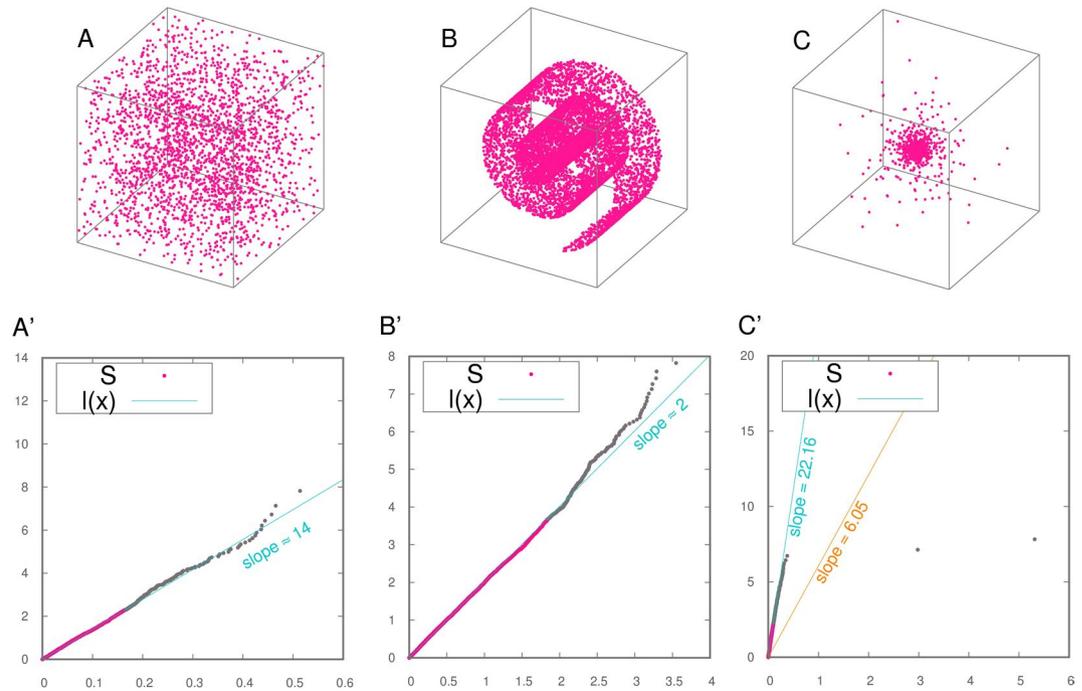

**Figure 1.** The fitting function $l(x)$ in three exemplar datasets of 2500 points. In the first column we display the dataset while in the second one we represent dataset $S$ (red dots) together with the discarded points (gray dots) and the fitting function $l(x)$. Panel A, A': cube in dimension 14 (in panel A only the first 3 coordinates are represented) analyzed with pbc. Panel B, B': a Swiss Roll. Panel C, C': a Cauchy dataset in dimension 20 (only the first 3 coordinates are represented).

Danco works marginally better in datasets characterized by the presence of sharp boundaries. Indeed such boundaries introduce an important violation to the assumption of local uniformity. In the Cauchy datasets TWO-NN achieves much better performances especially at high dimensions. On hypercubes without pbc and on Gaussians TWO-NN undergoes an overestimation, due to the presence of sharp boundaries, while for the same reason DANCo performs relatively well; adding pbcs allows TWO-NN to estimate correctly the dimension. In the case of Cauchy datasets TWO-NN slightly overestimates the ID due to the presence of outliers (in dimension 20 it gives an estimation of about 22), while DANCo meets significant difficulties (in dimension 20 it gives an estimation of about 13).

**Estimating a scale-dependent intrinsic dimension.** An important feature of the TWO-NN estimator is its locality: it provides an estimate of the ID by considering only the first and second neighbor of each point. This makes it suitable for analyzing how the ID varies with the scale, and distinguishing in this way the number of "soft" directions. As a basic example, consider a sample of points harvested form a uniform distribution on a plane perturbed by a Gaussian noise with variance $\sigma$ in a large number of orthogonal directions. This example mimics what is observed in samples extracted from a finite temperature molecular dynamics run, in which most of the possible directions are strongly disfavored by steric constraints. In the example, if the scale of interest is much larger than $\sigma$, say $10\sigma$ the relevant ID is 2.

We notice that what makes the notion of ID well defined in this example is the stability of the measure with respect to changes in the scale of interest: the ID would be 2 also on an even larger scale, say $100\sigma$.

In Nearest Neighbors-Based estimators the reference scale is the size of the neighborhood involved in the estimation; this depends on the density of points in the sample and does not necessarily coincide with the scale of interest. Going back to the example of the plane with noise, the more data points are used for the estimate, the smaller the average distance of the second neighbor will become, and the larger the ID. These observations suggest that in order to check the relevance of our measure we can study the stability of the estimation with respect to changes in the neighborhood size like in a standard block analysis. In the case of TWO-NN it is possible to modify the neighborhood size by reducing the number of points in the dataset: the smaller $N$, the larger the average distance to the second neighbor. In practice, similarly to the approach adopted in ref.[14], the analysis of the scaling of the dimension vs the number of points can be carried out by extracting subsamples of the dataset and monitoring the variation of the estimate $\hat{d}$ with respect to the number of points $N$. The relevant $ID$ of the dataset can be obtained by finding a range of $N$ for which $\hat{d}(N)$ is constant, and thus a plateau in the graph of $\hat{d}(N)$. The value of $d$ at the plateau is the number of "soft", or relevant, directions in the dataset.

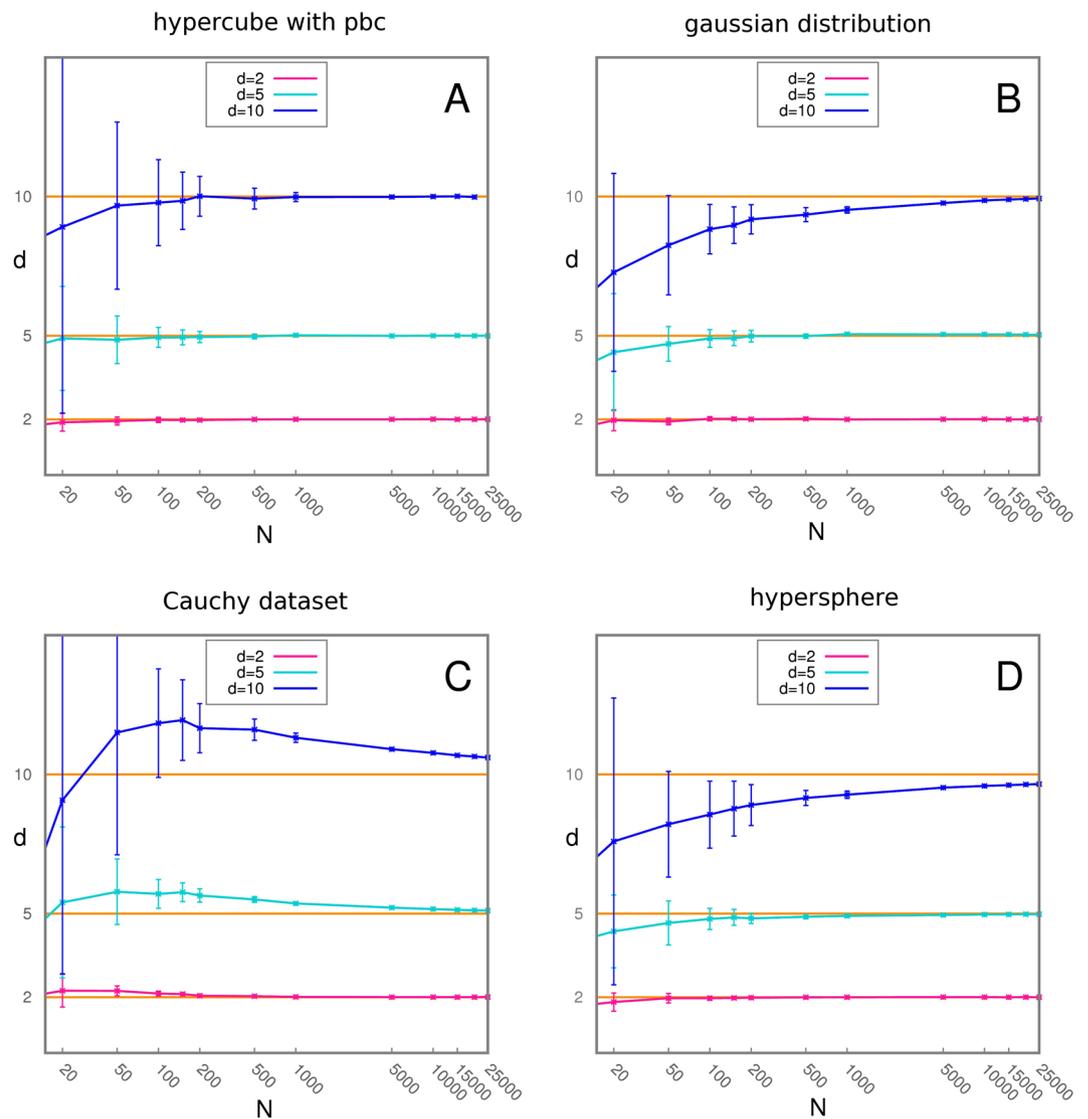

**Figure 2.** Scaling of the estimated ID with respect to the number of points; for each distribution and for a number of points going from 20 to 25000 we harvest 200 instances of the dataset and average the resulting estimates for the ID. The test is carried out in dimension 2, 5 and 10. Panel A: Hypercube with pbc. Panel B: gaussian distribution. Panel C: Cauchy dataset. Panel D: uniform distribution on a hypersphere.

In Fig. 3 we analyze the dimension. In Panel A we study the case of a uniform plane in dimension 2 perturbed by a high-dimensional gaussian noise with variance $\sigma$. We see that for $\sigma = 0.0001$ and $\sigma = 0.0002$ $d(N)$ displays a plateau around $N = 1000$, and the value of the dimension at the plateau is 2, equal to the number of soft directions in the dataset. As the number of points grows also noisy dimensions are sampled, and the value of the estimated ID increases. For critically low values of $N$ the estimated ID decreases to one, as expected (two points are always contained in a line). In Panel B we analyze a more challenging dataset composed of a two-dimensional Gaussian wrapped around a Swiss Roll and perturbed by a high-dimensional gaussian noise with variance $\sigma$. Also in this case we find a plateau, around 100 for $\sigma = 0.0002$ and around 500 for $\sigma = 0.0001$, at which the estimated dimension is 2. It is important to notice that even if the dataset analyzed in Panel B is far more complex than the simple plane in Panel A the behavior of the dimension vs the number of points is essentially the same in the two cases.

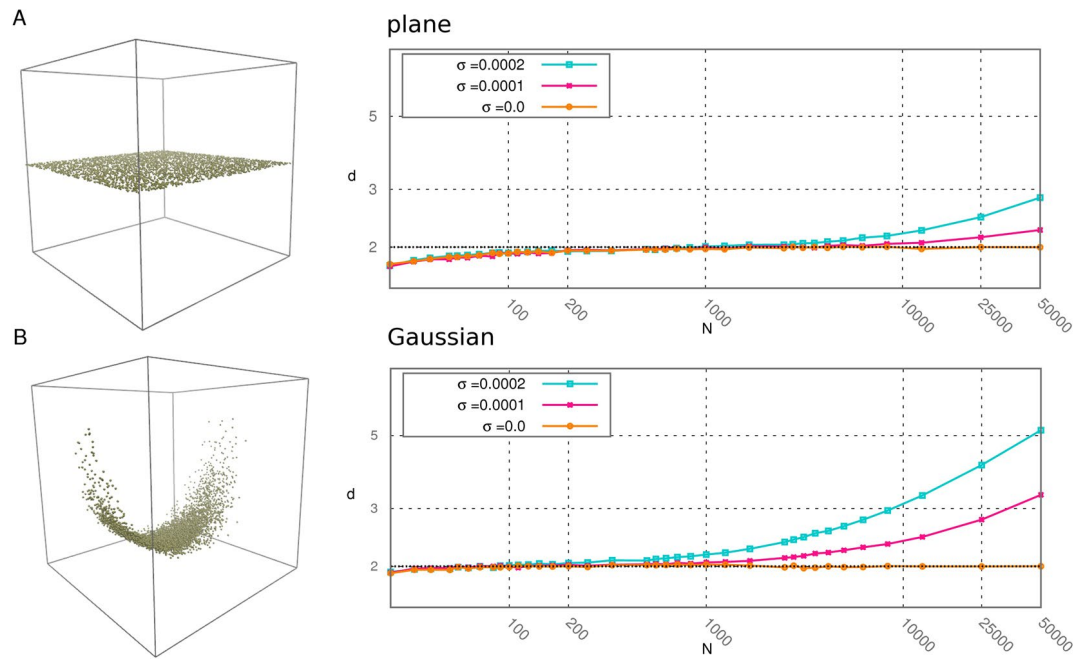

**Figure 3.** Estimated dimension *d* vs the number of points *N* in logarithmic scale; for each value of *N* the dataset is partitioned in a number of independent sets containing exactly *N* points, *d* is computed on each subdataset and a measure *d(N)* is obtained as an average of these values. In Panel A we study the case of a uniform plane of 50000 points in dimension 2 perturbed by a Gaussian noise with variance $\sigma$ along 20 independent directions; $\sigma$ takes the three values 0.0, 0.0001 and 0.0002. In Panel B we analyze a dataset composed of a two-dimensional Gaussian of 50000 points wrapped around a Swiss Roll and perturbed by a gaussian noise with variance $\sigma$ along 20 independent directions. Again $\sigma$ takes the three values 0.0, 0.0001 and 0.0002.

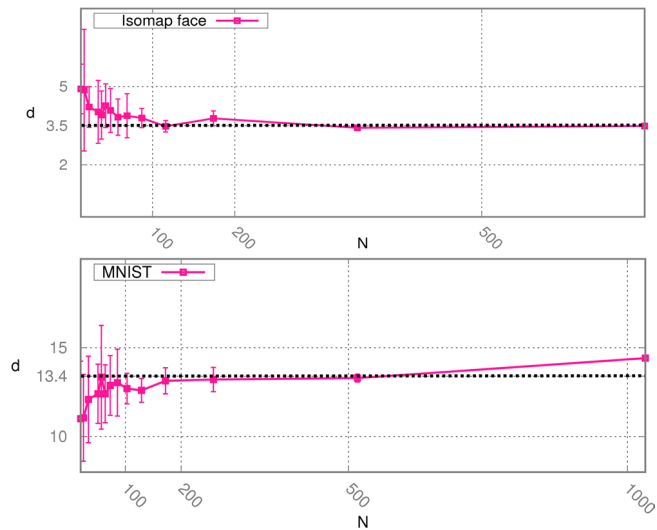

**Figure 4.** Scaling of the estimated ID with respect to the number of points for ISOMAP face (panel A) and MNIST database (panel B).

*Analysis of image datasets.* Estimating a scale-dependent intrinsic dimension is highly useful in case of real datasets. In Fig. 4 we compute the intrinsic dimension of two complex datasets: the Isomap face database and the handwritten "2"s from the MNIST database[9]. The first dataset consists of 598 vectors with 4096 components, representing the brightness values of 64 pixel by 64 pixel images of a face with different lighting directions and poses. The second one is composed by 1032 vectors with 784 components representing handwritten "2"s. Despite the relativey low number of points the block analysis is able to robustly detect the instrinsic dimension of the two datasets. In the case of Isomap faces we see a plateau for a number of points greater than roughly 400 and the measure of the ID in the range of the plateau is 3.5, slightly higher but consistent with 3, the value considered to be correct. In the case of MNIST dataset the plateau is located in a range between 300 and 500 points, and the measure of the ID corresponding to the plateau is 13.4, consistently with previous estimations that state the ID to be between 12 and 14[18,19]. In this case the measure we would obtain with the whole dataset would overestimate the ID due to the presence of noise, similarly to what observed in the artificial datasets in Fig. 3.

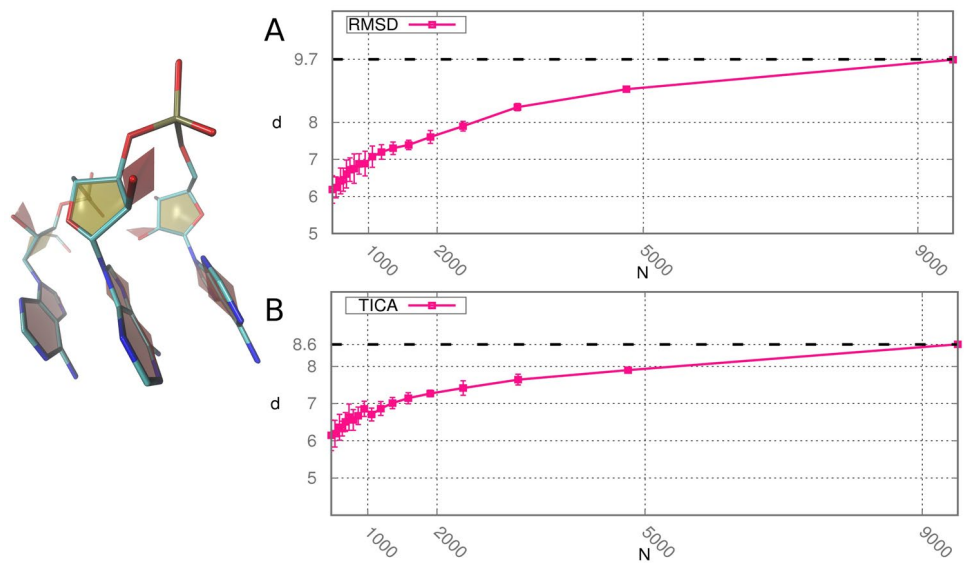

**Figure 5.** Scaling of the estimated ID with respect to the number of configurations for the dynamics of trinucleotide AAA in case of RMSD distances (panel A) and TICA distances (panel B). On the left a possible configuration is represented.

*Estimating the ID of trinucleotide AAA dynamics.* We finally estimate the ID of the configurational space explored during a molecular dynamics trajectory of the RNA trinucleotide AAA[17]. The dynamics was performed using GROMACS 4.6.7[20] at a temperature $T = 300$ K. RNA molecules were solvated in explicit water. From the original trajectory of 57 ms, we keep a configuration every 6 ns, obtaining a total number of 9512 structures. The simulation was originally carried out to provide insight into the main relaxation modes of short RNA. Computing the ID can provide a guideline for performing dimensionality reduction thus retaining in the description a meaningful number of variables. We perform the analysis of the ID making use of two notions of distance; the first one is the Euclidean distance between the coordinates associated to each sample by Time-lagged Independent Component Analysis (TICA)[21]. The second one is the Root Mean Square Deviation (RMSD) between all the atoms in the trinucleotide. These two distances are intrinsically different from each other, but strikingly the measure of the ID obtained in the two cases is comparable as shown in Fig. 5, with values of approximately 9.5 and 8.5 for the two metrics (estimated by using all the 9512 onfigurations). In the range of $N$ we considered, the estimate of $d$ slowly grows with a trend similar to the one observed in Fig. 2 on artificial data sets. It is possible in principle to further refine the procedure by fitting the these curves and finding the asymptotic value of $d$. Noticeably, the scaling features of $d$ vs $N$ with the two metrics are comparable and the ID values on the full datasets differ for only for one unit in dimension nine.

## Discussion

In this work we address the problem of finding the minimal number of variables needed to describe the relevant features of a dataset; this number is known as intrinsic dimension (ID). We develop TWO-NN, an ID estimator that employs only the distances to the first two nearest neighbors of every point. Considering a minimal neighborhood size has some some important advantages: first of all it allows to lower the effects of density inhomogeneities and curvature in the estimation process; moreover it grants a measure that does not mix the features of the dataset at different scales. In the case of locally uniform distributions of points TWO-NN relies on a robust theoretical framework while in the general case, namely in the presence of curvatures and density variations, TWO-NN is numerically consistent. In addition, it is able to provide reliable estimates even in the case of a low number of points.

A primary issue in the case of real datasets is discriminating the number of relevant dimensions. To this purpose we discuss a new method based on the use of TWO-NN to compute the ID on subsamples randomly extracted from the dataset, and analyze the behaviour of the estimated dimension with respect to the number of points. A plateau in such graph is indicative for a region in which the ID is well defined and not influenced by noise. The minimal neighborhood character of TWO-NN is a major advantage in this operation, since it allows to explore in a clean way the different length scales of the subsamples. We show that even in the case of a complex dataset displaying both curvature and density variations and perturbed by high dimensional gaussian noise we are able to successfully detect the number of relevant directions. We demonstrate that these features allow to estimate the ID even in real world datasets including sets of images and a set of configurations along a finite temperature molecular dynamics trajectory of a biomolecule in water solution. Finally we remark that using only two nearest neighbors grants a further advantage in terms of time complexity: by employing dedicated algorithms it is possible to find the first few neighbors of each point in an almost linearithmic time[22].

## References

1. Chen, M., Mao, S. & Liu, Y. Big data: a survey. *Mobile Networks and Applications* **19**, 171–209, https://doi.org/10.1007/s11036-013-0489-0 (2014).
2. Campadelli, P., Casiraghi, E., Ceruti, C. & Rozza, A. Intrinsic dimension estimation: Relevant techniques and a benchmark framework. *Mathematical Problems in Engineering* **2015**, https://doi.org/10.1155/2015/759567 (2015).
3. Piana, S. & Laio, A. Advillin folding takes place on a hypersurface of small dimensionality. *Phys. Rev. Lett.* **101**, 208101, https://doi.org/10.1103/PhysRevLett.101.208101 (2008).
4. Granata, D. & Carnevale, V. Accurate estimation of the intrinsic dimension using graph distances: Unraveling the geometric complexity of datasets. *Scientific Reports* **6**, https://doi.org/10.1038/srep31377 (2016).
5. Krueger, N. & Felsberg, M. A continuous formulation of intrinsic dimension. In *Proceedings of the British Machine Vision Conference*, 27.1–27.10, https://doi.org/10.5244/C.17.27 (BMVA Press, 2003).
6. Jolliffe, I. *Principal component analysis*, https://doi.org/10.1016/0169-7439(87)80084-9 (Wiley Online Library, 2002).
7. Cox, T. F. & Cox, M. A. *Multidimensional scaling*, https://doi.org/10.1007/978-3-540-33037-0_14 (CRC press, 2000).
8. Roweis, S. T. & Saul, L. K. Nonlinear dimensionality reduction by locally linear embedding. *Science* **290**, 2323–2326, https://doi.org/10.1126/science.290.5500.2323 (2000).
9. Tenenbaum, J. B., De Silva, V. & Langford, J. C. A global geometric framework for nonlinear dimensionality reduction. *science* **290**, 2319–2323, https://doi.org/10.1126/science.290.5500.2319 (2000).
10. Tribello, G. A., Ceriotti, M. & Parrinello, M. Using sketch-map coordinates to analyze and bias molecular dynamics simulations. *Proceedings of the National Academy of Sciences* **109**, 5196–5201, https://doi.org/10.1073/pnas.1201152109 (2012).
11. Grassberger, P. & Procaccia, I. Characterization of strange attractors. *Physical review letters* **50**, 346, https://doi.org/10.1103/PhysRevLett.50.346 (1983).
12. Kégl, B. Intrinsic dimension estimation using packing numbers. In *Advances in neural information processing systems* 681–688 (2002).
13. Fan, M., Qiao, H. & Zhang, B. Intrinsic dimension estimation of manifolds by incising balls. *Pattern Recognition* **42**, 780–787, https://doi.org/10.1016/j.patcog.2008.09.016 (2009).
14. Badii, R. & Politi, A. Hausdorff dimension and uniformity factor of strange attractors. *Physical review letters* **52**, 1661, https://doi.org/10.1103/PhysRevLett.52.1661 (1984).
15. Levina, E. & Bickel, P. J. Maximum likelihood estimation of intrinsic dimension. In *Advances in neural information processing systems* 777–784 (2004).
16. Ceruti, C. *et al.* Danco: An intrinsic dimensionality estimator exploiting angle and norm concentration. *Pattern recognition* **47**, 2569–2581, https://doi.org/10.1016/j.patcog.2014.02.013 (2014).
17. Pinamonti, G. *et al.* Predicting the kinetics of rna oligonucleotides using markov state models. *Journal of Chemical Theory and Computation* **13**, 926–934, https://doi.org/10.1021/acs.jctc.6b00982. PMID: 28001394 (2017).
18. Hein, M. & Audibert, J.-Y. Intrinsic dimensionality estimation of submanifolds in r d. In *Proceedings of the 22nd international conference on Machine learning* 289–296, https://doi.org/10.1145/1102351.1102388 (ACM, 2005).
19. Costa, J. A. & Hero III, A. O. Determining intrinsic dimension and entropy of high-dimensional shape spaces. In *Statistics and Analysis of Shapes* 231–252, https://doi.org/10.1007/0-8176-4481-4 (Springer, 2006).
20. Pronk, S. *et al.* Gromacs 4.5: a high-throughput and highly parallel open source molecular simulation toolkit. *Bioinformatics* btt055, https://doi.org/10.1093/bioinformatics/btt055 (2013).
21. Molgedey, L. & Schuster, H. G. Separation of a mixture of independent signals using time delayed correlations. *Phys. Rev. Lett.* **72**, 3634–3637, https://doi.org/10.1103/PhysRevLett.72.3634 (1994).
22. Muja, M. & Lowe, D. G. Scalable nearest neighbor algorithms for high dimensional data. *Pattern Analysis and Machine Intelligence, IEEE Transactions on* **36**, https://doi.org/10.1109/TPAMI.2014.2321376 (2014).

### Acknowledgements

We want to acknowledge Daniele Granata and Alex Rodriguez for their useful advice. We also acknowledge Michele Allegra, Giovanni Pinamonti and Antonietta Mira.

### Author Contributions

Laio, Facco, d'Errico and Rodriguez designed and performed the research. Laio and Facco wrote the manuscript text, prepared the figures and reviewed the manuscript.

### Additional Information

**Supplementary information** accompanies this paper at https://doi.org/10.1038/s41598-017-11873-y.

**Competing Interests:** The authors declare that they have no competing interests.

**Publisher's note:** Springer Nature remains neutral with regard to jurisdictional claims in published maps and institutional affiliations.